\documentclass[10pt,twocolumn,letterpaper]{article}

\usepackage{cvpr}              
\usepackage{enumitem}       
\usepackage{url}            
\usepackage{booktabs}       
\usepackage{amsfonts}       
\usepackage{nicefrac}       
\usepackage{microtype}      
\usepackage{xcolor}         
\usepackage{amsmath} 
\usepackage{multirow}
\usepackage{pifont}
\usepackage{fontawesome}
\usepackage{bbding}
\usepackage[numbers]{natbib}
\usepackage{colortbl}
\usepackage{graphicx}
\graphicspath{{images/}}
\newcommand{\fst}{\textbf}
\newcommand{\scd}{\underline}
\newcommand{\cmark}{\ding{51}}
\newcommand{\xmark}{\ding{55}}

\definecolor{cvprblue}{rgb}{0.21,0.49,0.74}
\usepackage[pagebackref,breaklinks,colorlinks,allcolors=cvprblue]{hyperref}

\def\paperID{31281} 
\def\confName{CVPR}
\def\confYear{2026}

\title{Proxy3D: Efficient 3D Representations for Vision-Language Models \\ via Semantic Clustering and Alignment}

\author{Jerry~Jiang\textsuperscript{\rm 1,*}
\quad~Haowen~Sun\textsuperscript{\rm 1,*}
\quad~Denis~Gudovskiy\textsuperscript{\rm 2}\\
\quad~Yohei~Nakata\textsuperscript{\rm 3}
\quad~Tomoyuki~Okuno\textsuperscript{\rm 3}
\quad~Kurt~Keutzer\textsuperscript{\rm 4}
\quad~Wenzhao~Zheng\textsuperscript{\rm 4,$\dagger$}\\
{\textsuperscript{\rm 1}Tsinghua University} ~~
{\textsuperscript{\rm 2}Panasonic AI Lab} ~~
{\textsuperscript{\rm 3}Panasonic DX-CPS} ~~
{\textsuperscript{\rm 4}UC Berkeley} ~~ \\
Project Page: \url{https://wzzheng.net/Proxy3D}\\
}

\renewcommand{\footnoterule}{%
  \kern -3pt
  \hrule width 0.9\linewidth height 0.4pt
  \kern 2.6pt
}

\begin{document}

\twocolumn[{
\renewcommand\twocolumn[1][]{#1}%
\maketitle
\vspace{-7mm}
\centering
\includegraphics[width=1\linewidth]{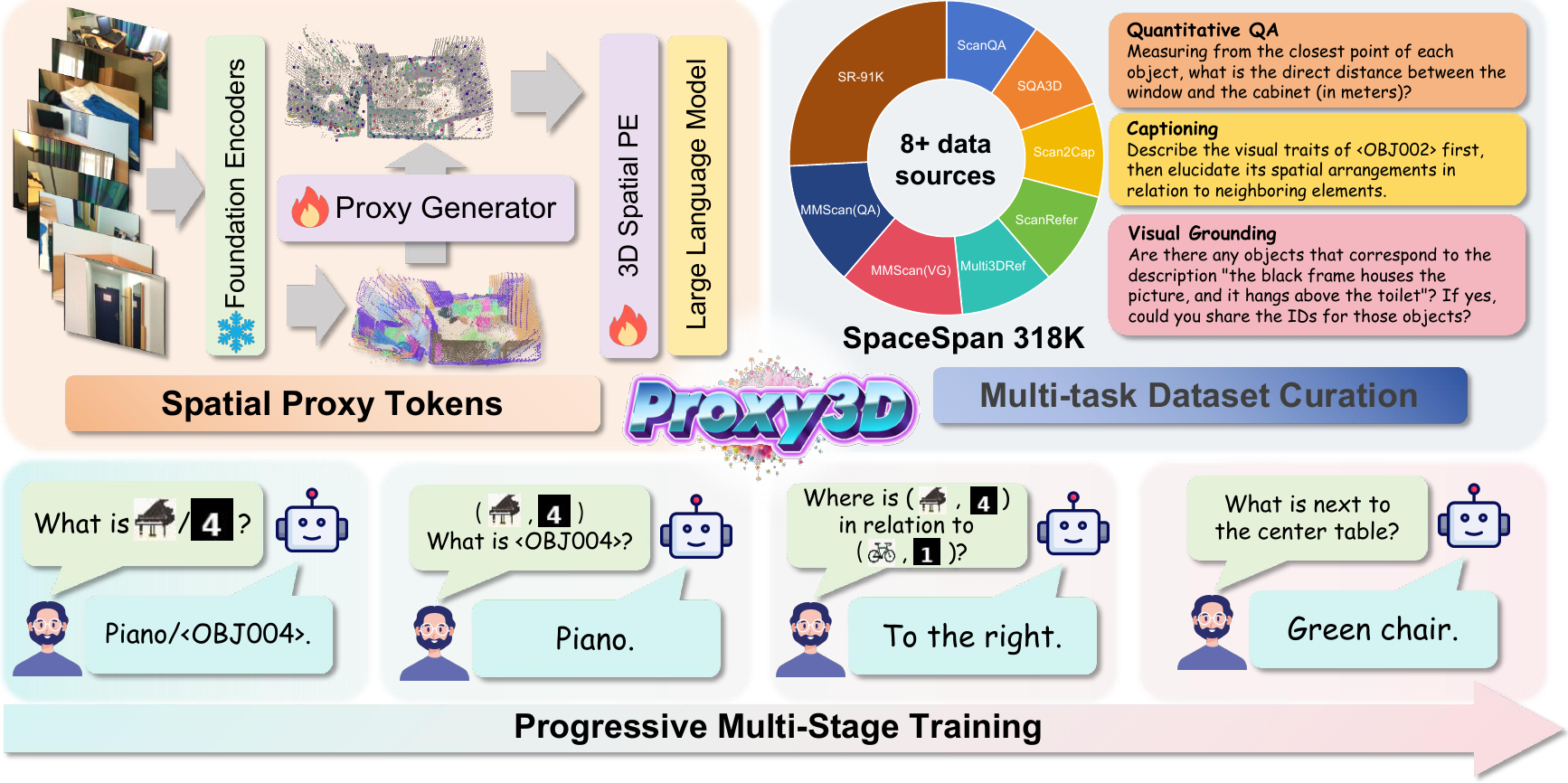}
\vspace{-8mm}
\captionof{figure}{\textbf{Overview of Proxy3D:} our 3D proxy representations are extracted from a set of pretrained encoders, their sequence length is compressed by the semantic-aware clustering followed by the multi-stage alignment with a language model using our SpaceSpan dataset.}
\vspace{3mm}
\label{fig:teaser}
}]

\begingroup
\renewcommand\thefootnote{}
\footnotetext{
  $^*$Equal contributions.
  $^\dagger$Corresponding author.
}
\endgroup

\begin{abstract}

Spatial intelligence in vision-language models (VLMs) attracts research interest with the practical demand to reason in the 3D world.
Despite promising results, most existing methods follow the conventional 2D pipeline in VLMs and use pixel-aligned representations for the vision modality.
However, correspondence-based models with implicit 3D scene understanding often fail to achieve spatial consistency, and representation-based models with 3D geometric priors lack efficiency in vision sequence serialization. 
To address this, we propose a Proxy3D method with compact yet comprehensive 3D proxy representations for the vision modality.
Given only video frames as input, we employ semantic and geometric encoders to extract scene features and then perform their semantic-aware clustering to obtain a set of proxies in the 3D space.
For representation alignment, we further curate the SpaceSpan dataset and apply multi-stage training to adopt the proposed 3D proxy representations with the VLM.
When using shorter sequences for vision information, our method achieves competitive or state-of-the-art performance in 3D visual question answering, visual grounding and general spatial intelligence benchmarks.
\end{abstract}
\section{Introduction}
\label{sec:intro}
\noindent
Spatial reasoning is a fundamental aspect of human intelligence~\citep{think}. When exploring a new scene, our vision senses 2D visual inputs as 3D spatial information and, further decoded into language modality, enables us to describe spatial relationships. Recent vision-language models (VLMs) and more general multimodal large language models (MLLMs) are equipped with a similar perception and, hypothetically, can achieve human-level spatial intelligence~\citep{cai2025}. 
At the same time, \citet{think} concludes that current MLLMs form a series of local world models in the vicinity of the ego image perspective, rather than a unified global model from a given video. Similarly, \citet{cai2025} quantitative evaluations show a large discrepancy between human-level spatial intelligence and the one in MLLMs. 

We argue that the \textit{representation of spatial information is crucial} for accurate and efficient reasoning in our 3D world. 
For example, VLMs with a correspondence objective achieve 3D spatial awareness implicitly by matching and aligning features across image frames~\citep{3DRS, llava-3D, spatialcot,  gpt4scene, liu2025coarse}. However, their learned representations suffer from inefficient usage of training data and spatial inconsistencies, resulting in a lack of global scene understanding and high computational costs.
Unlike it, representation-based methods explicitly model 3D scenes by leveraging 2D image features obtained from a pretrained vision encoder. Recent research have applied classic representations for 3D world modeling \eg, point clouds \citep{pointllm} \etc. Another line of research aims to develop a unified representation for 3D world \citep{SpatialMLLM, PQ3D, LEO-VL, VLM-3R}. However, the fundamental challenge remains: \textit{how to construct a sequence of tokens with accurate spatial information for a MLLM while minimizing its size?}

In this work, we find that the encoded vision modality has sparse semantic distribution and, therefore, we can leverage latent-space clustering to semantically compress 3D scenes. Our approach, dubbed \textit{Proxy3D}, with clustered proxy 3D features avoids complex neural network-based serializations as in \eg, \citet{Pointtransformer}. Lastly, we propose an iterative alignment to adopt our compressed 3D representations with a language model during the training phase using the \textit{SpaceSpan} dataset. Our main contributions are as follows:
\begin{itemize}[leftmargin=*]
\item We curate a 318K \textit{SpaceSpan} dataset with the unified data format that incorporates heterogeneous visual information.
\item We propose \textit{Proxy3D}, a method for aggregating compact yet comprehensive representations for spatial reasoning.
\item We introduce a multi-stage training pipeline that iteratively improves the MLLM's 3D scene understanding with the data-efficient representation alignment.
\end{itemize}
\section{Related Work}\label{gen_inst}

\textbf{Datasets and benchmarks for 3D scene understanding.} Main tasks for evaluating spatial intelligence in 3D-VLMs are QA, object grounding, and dense scene captioning. The former answers questions about spatial relationships between objects given visual input. For example, ScanQA \citep{Scannet} and SQA3D \citep{sqa3d}, from ScanNet \citep{Scannet}, are widely used for 3D QA evaluation.  Visual grounding (VG) identifies precise object locations using natural language queries. In the 2D domain, impressive and representative progress \citep{hu2016segmentation, yang2024language} has been made through referring image segmentation by utilizing datasets such as RefCOCO \citep{refcoco} and G-Ref \citep{gref}. More recently, common benchmarks for this task in the 3D space have been proposed, e.g., ScanRefer \citep{scanrefer} and Multi3DRefer \citep{multi3drefer}. Unlike VG, dense scene captioning (Scan2cap \citep{scan2cap}) estimates all object localizations and generates detailed descriptions. Other benchmarks \eg, VSI-Bench \cite{think} combine these tasks with spatiotemporal reasoning.

Although spatial intelligence is rapidly advancing with recent 3D-VLMs, its performance still falls short of understanding 3D scenes at the human level \citep{cai2025}. One of the reasons is the limited amount of training data, including the lack of 3D vision-language queries and object-object spatial relationship pairs for modality alignment. To overcome this, we propose a novel SpaceSpan dataset that is constructed on top of previous datasets but with more rich spatial reasoning-related queries and the unified data format. 

\textbf{3D scene representation modeling.} Recent 3D-VLMs, a subset of more general multimodal large language models (MLLMs), have been explored in several directions.  Correspondence-based methods assess video frame similarities to develop latent-space spatial cognition in LLMs. For example, 3DRS \citep{3DRS} develops 3D awareness using multi-view image correspondence with visual feature alignment, and Video-3D LLM \citep{video3dllm} achieves alignment using video frames via proposed 3D positional encoding. SR-3D \citep{SR-3D} extends global 3D positional embedding with canonical positional encoding. Ross3D \citep{ross3d} introduces explicit reconstruction between image views to inject 3D awareness. Despite promising results, these methods suffer from spatial inconsistencies and inefficient training data usage.

On the other hand, earlier research have explored various explicit representations to model 3D scenes \eg, point clouds \citep{pointllm, shapellm, ll3da}, depth maps \citep{spatialrgbt, spatialvlm}, 3DGS \citep{splattalk}, graphs \citep{3DGraphLLM}, and sparse spatiotemporal scene maps \citep{stsg}. However, they typically produce computationally-inefficient 3D scene representations with fixed geometric priors. More recent representation-based methods increase efficiency using \eg, point clouds with serialization for sequential transformer processing \citep{wang2023octformer, liu2023flatformer, Pointtransformer}. As a shortcoming, a na\"ive point cloud sequence cannot model the underlying complex spatial relationships using the cross-attention mechanism. To address this, recent Spatial-MLLM \citep{SpatialMLLM}, PQ3D \citep{PQ3D}, LLaVA-3D \citep{llava-3D}, LEO-VL \citep{LEO-VL} and VLM-3R \citep{VLM-3R} aim to develop a unified representation spanning geometric priors, instance-level visual features and global attributes. In our Proxy3D, we focus on preserving the benefits of representation-based methods, while learning compact 3D proxy representations to minimize MLLM's computational complexity.





\begin{figure*}[t]
\centering
\includegraphics[width=\textwidth]{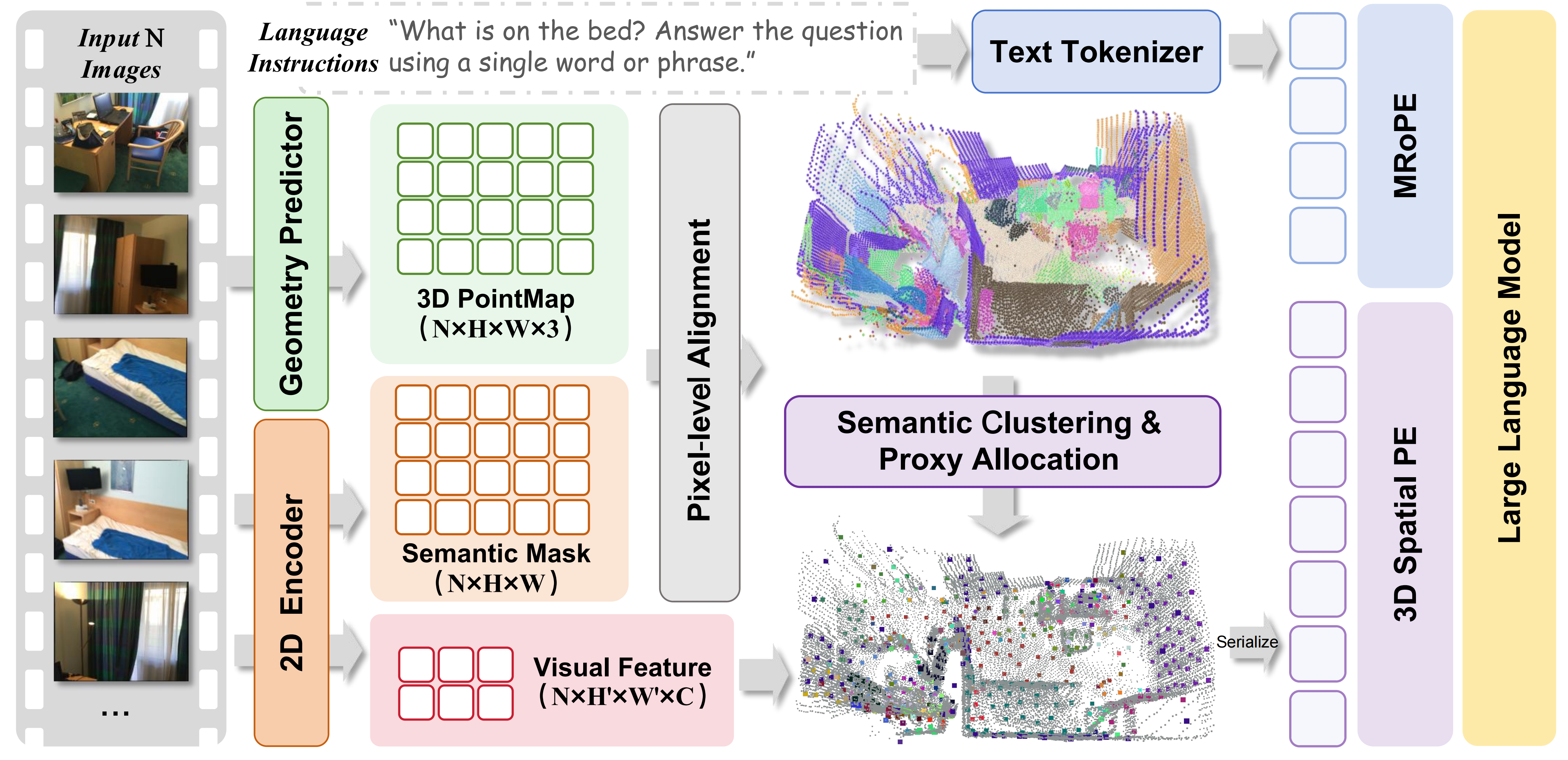}
\vspace{-7mm}
\caption{\textbf{Proxy3D architecture.} A geometry predictor and a semantic encoder output latent features of vision modality. Then, our proxy 3D representations are clustered to reduce complexity. Lastly, multi-stage training aligns proxy features with the language model.} 
\vspace{-5mm}
\label{fig:2}
\end{figure*}

\section{Proposed Method}\label{headings}
\subsection{Proxy3D Architecture}\label{sub:arch}


\paragraph{Feature extraction.} We employ spatial features from pretrained encoders with both semantic and spatial geometric information. First, $N$ RGB image frames $\{I_i\}_{i=1}^N$ with the $H\times W\times 3$ resolution are processed by a 2D visual encoder \citep{qwen25vl}. As a result, we obtain feature maps $\{F_i\}_{i=1}^N$, where the size of each $F_i\in \mathbb{R}^{H'\times W'\times C}$ depends on the encoder's latent dimension $C$ and the patch size $q$ that defines the downsampled height $H' = \lfloor H/q \rfloor$ and width $W' = \lfloor W/q \rfloor$.

Next, we use a geometry predictor \citep{vggt} to extract a set of point maps $\{P_i\}_{i=1}^N$ from image frames, where $P_i \in \mathbb{R}^{H \times W \times 3}$. To semantically group all features, we also apply a 2D segmentation model \citep{ravi2024sam2} and extract pixel-level segmentation masks $\{M_i\}_{i=1}^{N}$, where $M_i \in \mathbb{Z}^{H \times W}$. In order to align point maps and masks with the image features, we patchify them according to the selected patch size $q$ and produce the aligned sets $\{M_{j}, P_{j}\}_{j=1}^{L}$, where each element $M_{j} \in \mathbb{Z}^{q \times q}$, $P_{j}\in \mathbb{R}^{q \times q \times 3}$ and the sequence length $L=N\times H'\times W'$. To unify semantic information within each patch, we assign a $M_{j}$ label as the label of an object with the largest area and also normalize $P_j$ point map by that object area.

Then, we obtain triplets $\{F_j, P_j, M_j\}_{j=1}^{L}$ with aligned frame resolutions. Each element in the triplet defines either a spatial point in the space or a semantic group. For convenience, we flatten the patch-wise triplets along the latent dimension, yielding vectors $\{\mathbf{f}_j, \mathbf{p}_j, \mathbf{m}_j\}_{j=1}^{L}$ for the following semantic clustering.

\textbf{Semantic clustering.} To reduce the sequence length for computational efficiency, we propose to group the former triplets based on their semantic labels $g$ in the mask $\mathbf{m}_j$ as
\begin{equation}
\label{eq:gl}
\mathcal{G}_g = \{\mathbf{f}_j, \mathbf{p}_j \mid \mathbf{m}_j = g \},~\mathrm{and}~j=1, 2, \ldots, L,
\end{equation}
where $\mathcal{G}_g$ represents a semantically-aware set of features.

Inspired by \citet{PointGPT}, we cluster the group-aware sets in Equation (\ref{eq:gl}) using $K$-nearest neighbors (KNN) with a selected number of proxies $K_g$ for each semantic group as
\begin{equation}
\label{eq:knn}
\{\mathcal{C}_{g, j}\}_{j=1}^{K_g} = \mathrm{KNN} \left(\mathcal{G}_g, \mathbf{p}_k \right). 
\end{equation}

Using the result of Equation (\ref{eq:knn}), we can define the transformed visual features $\mathbf{z}_{g, j}$ and coordinates $\mathbf{c}_{g, j}$ that form our semantically-grouped set of 3D proxies as
\begin{equation}
\label{eq:proxy_definitions}
\mathcal{P} = \{ \mathbf{z}_{g, j}, \mathbf{c}_{g, j} \} = \{\mathbf{f}_j, \mathbf{p}_j \mid g \},~\mathrm{and}~ g,j \in \{\mathcal{C}_{g, j}\}.
\end{equation}

To further reference scene objects using their labels, we introduce identifier embeddings in Section~\ref{sub:training}.


\textbf{Proxy allocation.} We dynamically allocate $K_g$ based on each semantic group's proportion in the overall sequence \ie, $K_g \propto |\mathcal{G}_g|/L$. We assign an initial non-zero number of proxies to each group in case of empty groups to ensure that no instance is overlooked.

\textbf{Proxy3D sequence serialization.} We apply Breadth-first search (BFS) \citep{zhou2006breadth} traversal to our 3D group centers in Equation (\ref{eq:proxy_definitions}) and serialize the 3D visual embeddings into a list of scene tokens, starting from the root node of the closest 3D segment to the origin. Then, the segments that are spatially close to each other are also neighbors in the serialized sequence. This benefits an LLM to more accurately capture spatial relationships between various objects. 

\begin{figure*}[t]
\centering
\includegraphics[width=1\textwidth]{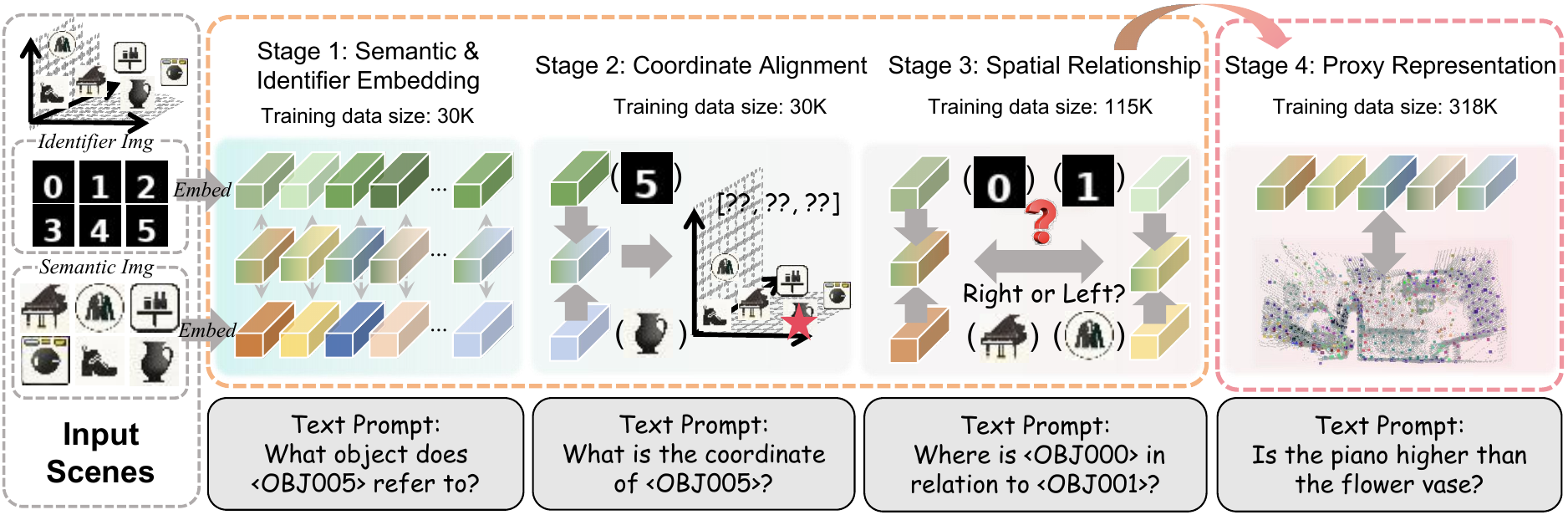}
\vspace{-7mm}
\caption{\textbf{Proxy3D multi-stage training.} Each stage in our progressive iterative training aims to develop a certain spatial intelligence skill from the easiest one to more complex ones: we begin with the simplified image-text alignment to actual images with spatial reasoning.}
\vspace{-5mm}
\label{fig:3}
\end{figure*}

\textbf{3D spatial position embeddings.} To further inject geometric priors, we apply 3D position embeddings to encode spatial information. Following \citet{LEO-VL}, we use rotary position embeddings (RoPE) \citep{su2024roformer} to the vertical position indices $\mathcal{H}$, and learnable Fourier embeddings \citep{li2021learnable} to the width and length $\mathcal{\{ W\times L \}}$. This can be expressed by
\begin{equation}
\mathbf{z}_{g, j}' = R\left(\mathbf{c}_{g, j \in \mathcal{H}} \right) \mathbf{z}_{g, j} + F\left(\mathbf{c}_{g, j\in \mathcal{\{ W\times L \}}} \right),
\end{equation}
where $R(\cdot)$ is the RoPE 2D rotation matrix and the Fourier embeddings $F(\cdot)$ are learned by an MLP.

The RoPE captures when objects move in the vertical dimension and the additive Fourier embeddings learn holistic spatial information for all objects. Note that multimodal RoPE is often applied to text tokens by VLMs \citep{qwen25vl}.

After applying the BFS and 3D positional information, our 3D proxy features can be written by
\begin{equation}
\label{eq:final_sequence}
\mathbf{Z} = [\mathbf{Z}_1, \mathbf{Z}_g, \ldots, \mathbf{Z}_G ],~\textrm{and}~ \mathbf{Z}_g = [ \mathbf{z}_{g,1}', \mathbf{z}_{g,2}', \ldots,\mathbf{z}_{g,K_g}' ],
\end{equation}
where the matrix $\mathbf{Z} \in \mathbb{R}^{K \times C}$ is a concatenation of variable-length matrices $\mathbf{Z}_g \in \mathbb{R}^{K_g \times C}$ for each semantic group $g=1\ldots G$ that have been sorted by the BFS and $K \ll L$.

\subsection{SpaceSpan Dataset and Multi-stage Training}\label{sub:training}

\textbf{SpaceSpan dataset.} We curate a 318K high-quality training set with the unified data format and the detailed description in Appendix. In short, it consists of 155K data points from the most common 3D datasets \citep{scanqa, sqa3d, scan2cap, scanrefer, multi3drefer}, and another 163K data points from recent MMScan \citep{mmscan} and SR-91K \citep{SpaceR}. In our multi-stage training scheme, we additionally apply a collection of 115K object-object relationship questions from MMScan to improve spatial reasoning skills.

\textbf{Object referencing.} MLLMs are mostly trained on 2D image inputs, achieving impressive performance in 2D feature interpretation. Though recent methods aggregate 2D inputs to form 3D features, MLLMs still struggle to interpret complex scenes using 2D embeddings. To address this, we propose spatial semantic positional embeddings as an intermediate representation that bridges spatial representations with LLM latent sequences.

We first generate a set of simplified identifier and semantic images as shown in Figure~\ref{fig:3} (left). 
We process these images through the vision encoder and obtain latent embeddings of size $1\times C$. In particular, we introduce two types of spatial embeddings that serve different purposes. First, the identifier embedding is an embedding-text pair that functions as a feature that unifies accurate object referencing with positional awareness. Second, the semantic embedding, derived by vision encoder from the simplified semantic symbol, serves as an efficient visual representation that describes a category of objects. We apply Stable Diffusion \cite{stablediffusion} to generate images of the latter semantic symbols and draw number characters to obtain the latter identifier images. 

Then, we can reference object categories by their semantic embeddings $\mathbf{f}_j^{sem} = G_{sem}(n_j)$, where $n_j$ denotes the category. Also, we can reference object instances by their identifier embeddings $\mathbf{f}_j^{id} = G_{id}(m_j)$, where $m_j$ denotes the object identifier.


Following Chat-Scene~\citep{chatscene} and MMScan~\citep{mmscan}, we define $m=100$ identifiers and $n=213$ object categories. We also express the text token $t_j$ that corresponds to an $m_j$-th identifier embedding $\mathbf{f}_j^{id}$ using the \texttt{<OBJXXX>} token format. Our approach is based on the rationale that models can effectively learn spatial relationships through simplified representations \eg, pieces in chess or stones in the Go game.

The proposed identifier embeddings also give us the advantage to reference objects without learnable embeddings as in \eg,~\citet{LEO-VL}. We extend visual prompting from 2D image to 3D feature space by directly injecting identifier embeddings into serialized proxy embeddings through additive fusion. To support our approach, we add the referenced identifier embeddings explicitly to the features in Equation (\ref{eq:gl}). Next, we describe the training with these embeddings.

\textbf{Multi-stage training.} In order to effectively form spatial understanding in MLLMs, we develop a progressive iterative training scheme as shown in Figure~\ref{fig:3}. 
In the first stage, we fuse the identifier and semantic embeddings for MLLM understanding, simulating the scene with simplified visual inputs. More specifically, the proxy embeddings in Equation (\ref{eq:final_sequence}) are substituted by the fused embeddings $\mathbf{f}^{sem}_j + \mathbf{f}^{id}_j$ generated using the corresponding object category $n_j$ and instance identifier $m_j$ for the $j$-th object. We replace all objects in the scene with such fused embeddings. An MLLM is then prompted with $t_j$ token to identify an object related to the given identifier \texttt{<OBJXXX>}.

\begin{figure}[t]
\centering
\includegraphics[width=1\columnwidth]{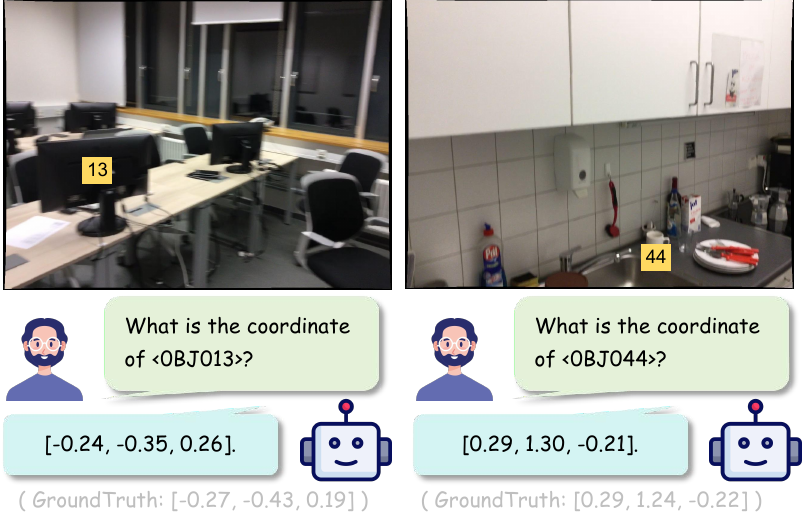}
\vspace{-7mm}
\caption{Coordinate alignment stage helps an MLLM to precisely align 3D positional embeddings with geometric coordinates.}
\vspace{-5mm}
\label{fig:4}
\end{figure}

In the second stage, coordinate alignment is carried out to train the 3D RoPE embeddings and develop the spatial size awareness for each identifier embedding. As shown in Figure~\ref{fig:4}, 3D position embedding is trained effectively with high accuracies on coordinates determination. With accurate knowledge of these simplified embeddings, MLLM is ready to explore space in the third stage, where we explicitly train MLLM to understand spatial relationships and effective positional encoding. In this stage, 115K data points have been collected from the object-object attribute slice of MMScan~\citep{mmscan} dataset. 

In the final stage, actual 3D scene proxies are used as visual inputs. When trained with the full 318K SpaceSpan dataset, MLLM shifts its knowledge from simplified visual inputs to real scene inputs.

\textbf{Instruction tuning objective.} We minimize the negative $\log$-likelihood loss for the autoregressive model with $\mathbf{\theta}$ parameters and 3D proxy representation in Equation (\ref{eq:final_sequence}) expressed by
\begin{equation}
\label{eq:sft}
\mathcal{L}(\mathbf{\theta}) = -\sum\nolimits^{r}_{i=K+1} \log P_{\mathbf{\theta}} (t_i | t_{<i}, \mathbf{Z}), 
\end{equation}
where $r$ is the response sequence length, $t_i$ is the $i$-th output token, $t_{<i}$ are the previous $i-1$ text tokens and $\mathbf{Z}$ is the introduced 3D proxy sequence.

\section{Experiments}\label{others}
\subsection{Experimental Setup}

\textbf{Implementation details.} We apply supervised multi-stage finetuning using (\ref{eq:sft}) objective for the pretrained Qwen2.5-VL-7B \citep{qwen25vl}. During the baseline finetuning, we set $K=450$ for the Proxy3D sequence length to balance scene details and training time. We apply $N=32$ uniformly sampled images with resolution $H=W=512$ for each scene as input frames. We apply VGGT \citep{vggt} as our geometry predictor, and SAM 2 \citep{ravi2024sam2} for 2D segmentation. The resolution of the latent embedding images is $H'=W'=28$. Note that the current VGGT only provides normalized point maps and, in order to preserve scale information, we estimate and apply scale factors to VGGT point maps using a procedure described in supplementary materials. The estimated time for each training stage is shown in Table~\ref{training time} when using $8\times$ A6000 NVIDIA GPUs and the proposed SpaceSpan dataset. We provide detailed hyperparameters in Appendix. 
\begin{table}[t]
\caption{Estimated Proxy3D training time in hours using Section~\ref{sub:training} training procedure and $8\times$ A6000 NVIDIA GPUs.}
\vspace{-3mm}
\label{training time}
\centering
\resizebox{0.65\linewidth}{!}{
\begin{tabular}{@{} *{4}{c} }
\toprule
\textbf{Stage 1} & \textbf{Stage 2} & \textbf{Stage 3} & \textbf{Stage 4}\\
\midrule
2 & 2 & 3 & 55\\
\bottomrule
\end{tabular}
}
\vspace{-5mm}
\end{table}

\textbf{Evaluation setup.} We compare Proxy3D with open-source correspondence- and representation-based 3D-VLMs from Section~\ref{gen_inst} as well as task-specific and proprietary baselines. Our comprehensive evaluation setup includes 3D question answering (QA) with ScanQA \citep{Scannet} and SQA3D \citep{sqa3d} benchmarks, visual grounding (VG) with ScanRefer \citep{scanrefer} and Multi3DRefer \citep{multi3drefer}, dense captioning (DC) using Scan2cap \citep{scan2cap} benchmark and VSI-Bench \cite{think} for spatial reasoning.

For 3D QA and DC benchmarks, we abbreviate performance metrics as "C" for CIDEr, "B-4" for BLEU-4, "M" for METEOR, "R" for ROUGE, and "EM" for top-1 exact match accuracy. 
In 3D VG, we report unique accuracy ("Uni"), overall accuracy ("Acc") and F$_1$ scores. 
Recent VSI-Bench for spatial reasoning contains 5,000 question-answer pairs with eight spatial tasks, including multiple-choice and numerical answers. We follow the VSI-Bench metric design and compute the mean exact accuracy for multiple-choice answers and the mean relative accuracy across confidence thresholds $\mathcal{C}=\{0.5,0.55,...,0.95\}$ for numerical answers.

\subsection{Quantitative Results}

\begin{table*}[t]
\caption{\textbf{Evaluation of 3D question answering, visual grounding and dense captioning.}  We follow the standard evaluation methodology for all benchmarks. We categorize models by their type, used vision modalities (P - point clouds, I - images, B - bird's-eye-view map, D - depth), sequence length $L$ (\# of tokens). The \fst{best} and the \scd{second best} results are highlighted. Our Proxy3D with Qwen2.5-VL backbone shows competitive or state-of-the-art results with the shortest sequence lengths. "‡" means usage of extra information from point clouds.}
\vspace{-3mm}
\label{QAVG_results}
\centering
\setlength{\tabcolsep}{2pt}
\resizebox{1.0\linewidth}{!}{
\begin{tabular}{@{} l c c *{2}{c} *{2}{c} *{2}{c} *{2}{c} *{2}{c} @{}}
\toprule
\multirow{2}{*}{Models} & \multirow{1}{*}{Vision} & \multirow{1}{*}{\# of} & \multicolumn{2}{c}{Scan2Cap (val)} & \multicolumn{3}{c}{ScanRefer (val)} & \multicolumn{2}{c}{Multi3DRefer (val)} & \multicolumn{2}{c}{ScanQA (val)} & SQA3D (val)\\
\cmidrule(lr){4-5} \cmidrule(lr){6-8} \cmidrule(lr){9-10} \cmidrule(lr){11-12} \cmidrule(lr){13-13}
& modal. & tokens & C@0.5 & B-4@0.5 & Uni@0.5 & Acc@0.25 & Acc@0.5 & F$_1$@0.25 & F$_1$@0.5 & C & EM & EM\\
\midrule
\multicolumn{9}{@{}l}{\emph{Task‐specific:}}\\
ScanQA~\citep{scanqa}     & P & 256 &  –  &   – & –  &  –  &   –  &    –  &   –  &  64.9 & 21.1 & 47.2 \\
Scan2Cap~\citep{scan2cap}        & P & 256 & 39.1 & 23.3 & – & – & – & –  & – & – & – & –\\
ScanRefer~\citep{scanrefer}      & P & 256 &   –  &   – & 53.5  &   37.3  &   24.3  &   –  & –  & – & – & –\\
\midrule
\multicolumn{9}{@{}l}{\emph{3D LLM-based:}}\\
LEO~\citep{leo}         & P,I & - &   –  & 72.4 & – & 38.2 & – & – & – & 80.0 & 24.5 & 50.0\\
Chat‐Scene~\citep{chatscene}  & P & - & 77.1 & 36.3 & 82.5 & 55.5 & 50.2 & 57.1 & 52.4 & 87.7 & 21.6 & 54.6 \\
Descrip3D~\citep{Descrip3D}    & I & - & 77.2 & 34.5 & \scd{83.2} & 57.2 & 51.8 & 59.4 & \scd{55.1} & 93.7 & 22.67 & 55.7\\
\midrule
\multicolumn{9}{@{}l}{\emph{Correspondence-based:}}\\
Qwen2‐VL‐7B ~\citep{qwen2vl}     & I & 8000 & 0 & 3.8 & 6.3 & 5.4 & 5.1 & 21.1 & 19.9 & 53.9 & 19.0 & 40.7\\
\rowcolor{gray!20} \textcolor{gray}{GPT4Scene (HDM)$^\textrm{‡}$} ~\citep{gpt4scene}         & B,I & 8000 & \textcolor{gray}{86.3} & \textcolor{gray}{40.6} & \textcolor{gray}{83.7} & \textcolor{gray}{62.6} & \textcolor{gray}{57.0} & \textcolor{gray}{64.5} & \textcolor{gray}{59.8} & \textcolor{gray}{90.9} & \textcolor{gray}{28.2} & \textcolor{gray}{57.4}\\
Video-3D-LLM~\citep{video3dllm}  & I & 8000 & \scd{83.8} & \scd{41.4} & 78.3 & 58.1 & 51.7 & 58.0 & 52.7 & \scd{102.1} & \scd{30.1} & 58.6 \\
Spatial-MLLM-4B \cite{SpatialMLLM} & I &   3100  &   –  & – &   –  &   –  &   –  &   –   &   –   & 91.8 & 26.3 & 55.9\\
3DRS~\citep{3DRS}                & I & 8000 & \fst{86.1} & \fst{41.6} & 77.9 & \fst{62.9} & \fst{56.1} &  \scd{60.4} & 54.9  & \fst{104.8} & \fst{30.3} & \scd{60.6}\\
\midrule
\multicolumn{9}{@{}l}{\emph{Representation-based:}}\\
LLaVA-3D~\citep{llava-3D}        & D,I  & 3096 & 79.2 & 41.1 & – & 54.1\tiny & 42.4 & – & – & 91.7 & 27.0 & 55.6 \\
LEO-VL~\citep{LEO-VL}         & D,I  & \scd{750} & –  &   – & –  &  –  &   –  &    –  &   –  & 100.4 & 22.6 & \fst{60.8} \\
\rowcolor{blue!15}
Proxy3D$_\textrm{Qwen2.5-VL-7B}$    & I & \fst{700} & 73.3 & 34.7 & \textbf{84.0} & \scd{59.6} & \scd{54.1} & \textbf{62.0} & \fst{57.5} & 93.6 & 25.2 & 57.5\\
\bottomrule
\end{tabular}
}
\end{table*}

\begin{table*}[!t]
\centering
\captionof{table}{\textbf{Evaluation of 3D spatial reasoning on VSI-Bench}. We use 16 frames as input for Qwen2.5VL-based baselines and, following the VSI-Bench setup, other open-source and proprietary models use from 16 to 32 image frames. The \fst{best} and the \scd{second best} results for open-source models are highlighted. Our Proxy3D with Qwen2.5-VL-7B backbone shows overall the second best result. At the same time, the gap with the human-level performance remains significant in spatial reasoning certain tasks. "‡" indicates tasks not specifically trained.} 
\vspace{-3mm}
\setlength\tabcolsep{3pt}
\resizebox{\textwidth}{!}{
\begin{tabular}{lcccccccccc}
\toprule
\multirow{2}{*}{Models}  & \multicolumn{4}{c}{Numerical Answer} & \multicolumn{4}{c}{Multiple-Choice Answer} & \multirow{2}{*}{Avg.} &\multirow{2}{*}{Rank} \\
\cmidrule(lr){2-5}\cmidrule(lr){6-9}
& Obj. Cnt. & Abs. Dist. & Obj. Size & Room Size & Rel. Dist. & Rel. Dir. & Route Plan$^\textrm{‡}$ & Appr. Order$^\textrm{‡}$ & &\\
\midrule
Human level & 94.3 & 47.0 & 60.4 & 45.9 & 94.7 & 95.8 & 95.8 & 100.0 & 79.2 & - \\
\midrule
\multicolumn{1}{l}{\textcolor{black}{\textit{Proprietary via API:}}} & & & & & & & & & &\\
GPT-4o~\cite{gpt4o}  & 46.2 & 5.3 & 43.8 & 38.2 & 37.0 & 41.3 & 31.5 & 28.5 & 34.0 & 8\\
Gemini-1.5 Pro~\cite{gemini} & 56.2 & 30.9 & 64.1 & 43.6 & 51.3 & 46.3 & 36.0 & 34.6 & 45.4 & 3\\
\midrule
\multicolumn{1}{l}{\textcolor{black}{\textit{Open-source:}}} & & & & & & & & & \\
InternVL2-40B~\cite{internvl}  & 34.9 & 26.9 & 46.5 & 31.8 & 42.1 & 32.2 & \scd{34.0} & 39.6 & 36.0 & 7\\
LLaVA-OV-72B~\cite{llavaov}  & 43.5 & 23.9 & 57.6 & 37.5 & \scd{42.5} & 39.9 & 32.5 & 44.6 & 40.2 & 5\\
LLaVA-Video-72B~\cite{videoit}  & 48.9 & 22.8 & 57.4 & 35.3 & 42.4 & 36.7 & \textbf{35.0} & \textbf{48.6} & 40.9 & 4\\
Qwen2.5-VL-7B~\cite{qwen25vl}  & 40.9 & 14.8 & 43.4 & 10.7 & 38.6 & 38.5 & 33.0 & 29.8 & 33.0 & 9\\
Qwen2.5-VL-72B~\cite{qwen25vl} & 25.1 & 29.3 & 54.5 & 38.8 & 38.2 & 37.0 & \scd{34.0} & 28.9 & 37.0 & 6\\
{Spatial-MLLM-4B}~\cite{SpatialMLLM}  & \fst{65.3} & \scd{34.8} & \scd{63.1} & \fst{45.1} & 41.3 & \scd{46.2} & 33.5 & \scd{46.3} & \fst{48.4} & \fst{1}\\
\rowcolor{blue!15}
Proxy3D$_\textrm{Qwen2.5-VL-7B}$  & \scd{63.9} & \textbf{41.9} & \textbf{67.2} & \scd{42.8} & \textbf{50.3} & \textbf{46.5} & 31.4 & 32.0 & \scd{47.0} & \scd{2}\\
\bottomrule
\end{tabular}
}
\vspace{-5mm}
\label{tab:vsibench}
\end{table*}

\begin{figure*}[t]
\centering
\includegraphics[width=1\linewidth]{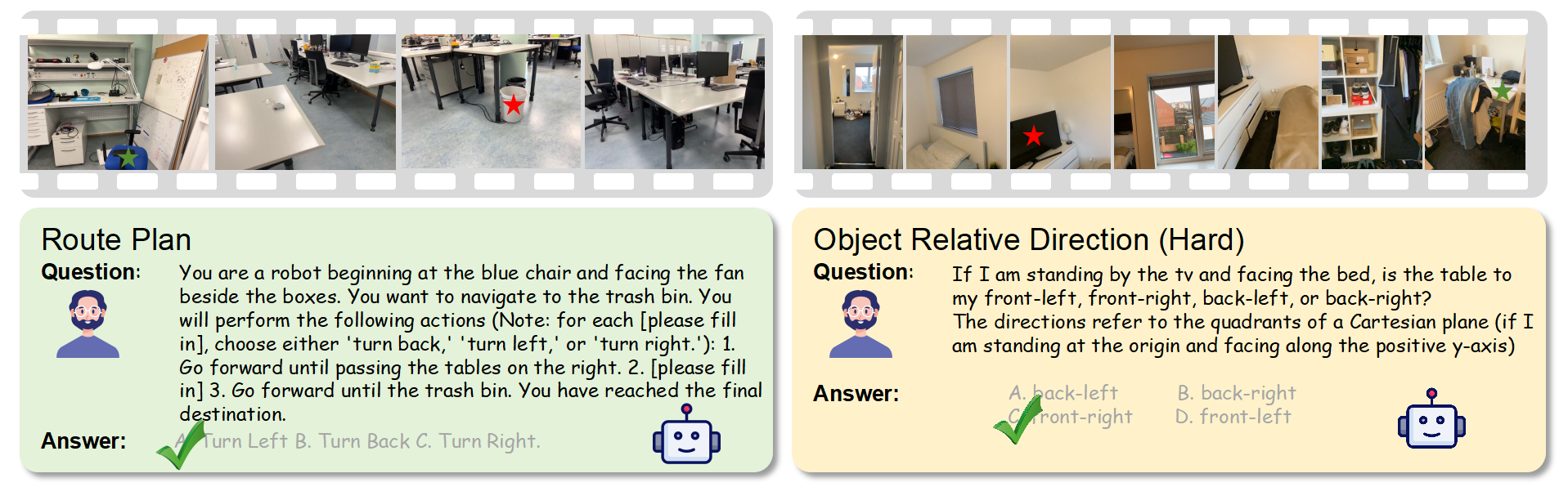}
\vspace{-7mm}
\caption{Proxy3D performance on VSI-Bench~\citep{think}. Left is on Scannet++~\citep{scannet++}, right is on ARKitScenes~\citep{arkitscenes}. Proxy3D generalizes well on unseen scenes, and is capable of solving difficult questions.}
\vspace{-5mm}
\label{fig:visualization}
\end{figure*}

\textbf{3D question answering.} We present ScanQA and SQA3D results in Table~\ref{QAVG_results}. Compared to correspondence-based models, Proxy3D achieves similar performance metrics with less than 10\% of visual tokens. The same conclusion is applicable to representation-based LLaVA-3D~\citep{llava-3D} with 3,096 tokens. Compared to LEO-VL~\citep{LEO-VL} with 750 tokens and an additional post-training SceneDPO objective, Proxy3D with only 700 tokens demonstrates nearly identical performance when benchmark results are available. This is justified because LEO-VL and Proxy3D share many architectural components, where the latter focuses on the sequence compression using semantic-aware clustering, and the former utilizes an extra training stage with positive and negative answer contrasting.

\textbf{3D visual grounding and dense captioning.} Proxy3D achieves state-of-the-art or second best results on ScanRefer and Multi3DRefer benchmarks, which can be attributed to accurate distinction of objects with the proposed semantic grouping. In particular, Proxy3D outperforms object-proposal methods \eg, Chat-Scene~\citep{chatscene} and Descrip3D~\citep{Descrip3D}. The latter models simply combine embeddings of each object into a sequence. In contrast, Proxy3D not only concatenates a series of object instances, but also performs semantic-aware grouping and compression. At the same time, the correspondence-based 3DRS~\citep{3DRS} is also competitive in these benchmarks but, again, it suffers from more than $10\times$ longer sequence length. Scan2Cap dense captioning is the toughest benchmark for all representation-based models that significantly underperform compared to the correspondence-based ones, possibly a tradeoff between simplicity and semantics.


\textbf{Spatial reasoning.} We present quantitative results for the VSI-Bench~\citep{think} in Table~\ref{tab:vsibench}. We compare Proxy3D to proprietary GPT-4o~\citep{gpt4o} and Gemini-1.5 Pro~\citep{gemini} baselines as well as open-source models: InternVL2~\citep{internvl}, LLaVA-OneVision~\citep{llavaov}, LLaVA-Video~\citep{videoit}, Spatial-MLLM~\citep{SpatialMLLM} and various variants of Qwen2.5VL~\citep{qwen25vl}. Surprisingly, only the Spatial-MLLM~\citep{SpatialMLLM} baseline has a marginal improvement over the proposed Proxy3D. Analysis of Spatial-MLLM work shows that, similar to LEO-VL \citep{LEO-VL}, it relies on post-training reward learning using group relative policy optimization (GRPO)~\citep{shao2024deepseekmath}. Therefore, we conclude that the reward learning step could be useful to further improve Proxy3D results. At the same time, Spatial-MLLM without sequence compression employs approximately $7\times$ more tokens (3,096 vs. 450) than in our Proxy3D.

\begin{figure}[!t]
\centering
\includegraphics[width=0.95\columnwidth]{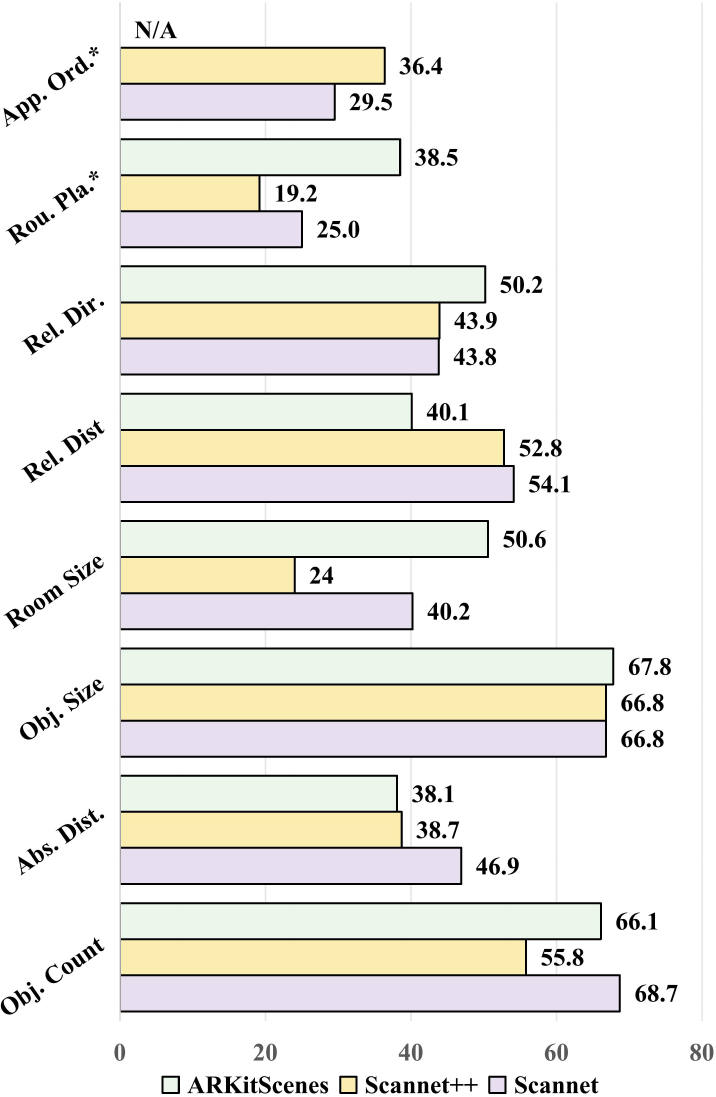}
\vspace{-3mm}
\caption{Comparison of VSI-Bench tasks and splits \ie ARKitScenes~\citep{arkitscenes}, Scannet++~\citep{scannet++} and Scannet~\citep{Scannet}. Results show Proxy3D robustness to data splits and uneven metrics across tasks.}
\label{fig:6}
\vspace{-5mm}
\end{figure}

Table~\ref{tab:vsibench} metrics for all models still have a large gap compared to the human level of spatial reasoning. We check if this is related to domain shifts in VSI-Bench heterogeneous dataset splits. Figure~\ref{fig:6} compares Proxy3D performance on all VSI-Bench tasks and data splits \ie ARKitScenes, Scannet++ and Scannet. Our study shows mostly uniform results across data splits, but large discrepancies between the type of spatial reasoning tasks. For example, object counting and size measuring is substantially closer or even exceed human level performance, but all models significantly lag behind in appearance order and route planning tasks. Visualization results are provided in Figure~\ref{fig:visualization}.

\begin{table*}[t] 
\caption{Ablation study on various aspects of the Proxy3D approach: inter-frame cross attention in vision encoder, semantic grouping, coordinate alignment, feature map resolution and number of proxy tokens. In this study, we justify effectiveness of the proposed methods in Proxy3D and present hyperparameters (feature map resolution and visual \# of tokens) for complexity-accuracy trade-off tuning.}
\vspace{-3mm}
\label{ablation}
\centering
\resizebox{1\linewidth}{!}{
\begin{tabular}{@{} c c c c c *{4}{c} *{2}{c} *{2}{c} @{}}
\toprule
\multirow{1}{*}{Inter-frame} & \multirow{1}{*}{Semantic} & \multirow{1}{*}{Coord.} & \multirow{1}{*}{Feature} & \multirow{1}{*}{\# of} & \multicolumn{4}{c}{ScanQA (val)} & \multicolumn{2}{c}{SQA3D (val)} & \multicolumn{2}{c}{ScanRefer (val)} \\
\cmidrule(lr){6-9} \cmidrule(lr){10-11} \cmidrule(lr){12-13}
cross attn. & grouping & align. &  map res. & tokens & EM &  B-4 & M &  C & EM & EM-R & Uni@0.5 & Acc@0.5 \\
\midrule
\cmark & \xmark & \cmark & 16x21 & 450 & 24.7 & 14.5 & 18.2 & 92.2 & 56.8 & 60.0 & 57.0 & 31.0\\
\cmark & \cmark & \cmark & 16x21 & 450 & 25.3 & 15.3 & 18.6 & 92.7 & 57.3 & 60.6 & 82.7 & 52.6 \\
\xmark & \cmark & \cmark & 32x42 & 700 & 25.4 & 15.3 & 18.6 & 93.1 & 56.9 & 60.2 & 83.2 & 53.8\\
\cmark & \cmark & \xmark & 32x42 & 700 & 24.9 & \textbf{15.7} & 18.7 & 93.4 & 56.7 & 59.8 & 83.6 & 53.8\\
\cmark & \cmark & \cmark & 32x42 & 700 & 25.2 & 15.4 & 18.7 & 93.6 & \textbf{57.5} & \textbf{60.8} & 84.0 & \textbf{54.1} \\
\cmark & \cmark & \cmark & 32x42 & 1000 & \textbf{25.6} & 15.5 & \textbf{18.8} & \textbf{94.3} & 57.0 & 60.2 & \textbf{84.7} & 53.8 \\
\bottomrule
\end{tabular}
\vspace{-3mm}
}
\end{table*}

\begin{figure*}[t]
\centering
\includegraphics[width=1\textwidth]{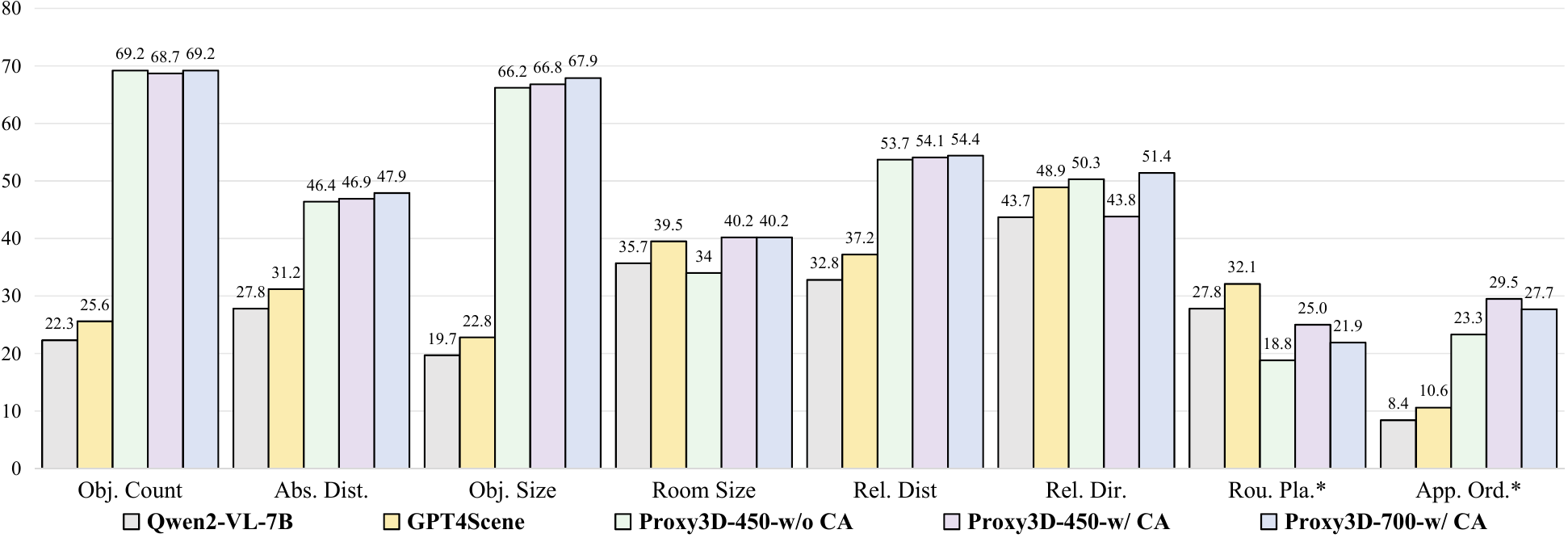}
\vspace{-7mm}
\caption{Ablation study on VSI-Bench's Scannet~\citep{Scannet} split. Proxy3D outperforms the base Qwen2-VL-7B and GPT4Scene by a large margin in object counting, size and distance estimation. Coordinate alignment (CA) and longer sequences further increase metrics.}
\label{fig:vsi-comparison}
\vspace{-3mm}
\end{figure*}

\subsection{Ablation Study}
\textbf{Feature map resolution and 3D proxy sequence length.} Table~\ref{ablation} compares various latent feature map resolutions and proxy sequence lengths while keeping the input image frame resolution constant. In particular, we apply bilinear interpolation to upsample feature maps to higher resolution. Results show that the higher resolution ($16\times21$ \vs $32\times42$) and sequences with larger length ($K=450, 700, 1000$)~lead to higher performance metrics. On the other hand, such scaling introduces additional computational overhead and the desired complexity-accuracy balance can be reached by tweaking the above model hyperparameters.

\begin{table}[t]\small 
\caption{Ablation study on dynamic allocation of group-aware proxies from Section~\ref{sub:arch} for Proxy3D with 700 tokens.} 
\vspace{-3mm}
\label{adaptive_alloc}
\centering
\setlength{\tabcolsep}{15pt}
\begin{tabular}{@{} c *{4}{c} }
\toprule
\multirow{1}{*}{\# of} & \multicolumn{4}{c}{Scan2Cap (val)} \\
\cmidrule(lr){2-5}
proxies & B-4 & M & R & C\\
\midrule
2 & 34.7 & 27.1 & 54.8 & 73.3\\
5 & \fst{34.9} & \fst{27.3} & \fst{55.1} & \fst{74.9}\\
10 & 34.4 & 27.1 & 54.7 & 73.2\\
\bottomrule
\end{tabular}
\vspace{-5mm}
\end{table}

\textbf{Coordinate alignment.} Coordinate alignment introduces significant improvement for accurate room size estimation, route planning and appearance order tasks as shown in Figure~\ref{fig:vsi-comparison} VSI-Bench ablation study. A qualitative example of coordinate alignment is also illustrated in Figure~\ref{fig:4}.

\textbf{Semantic grouping.} Table~\ref{ablation} ablates our semantic grouping. Results show moderate improvement in question answering benchmarks and significant impact on visual grounding task with more than 20 points increase in the overall accuracy. We conclude that na\"ive clustering without semantic grouping leads to inaccurate object referencing.

\textbf{Inter-frame attention.} We conduct an ablation study in Table~\ref{ablation} on a role of the inter-frame cross attention within a vision encoder. Proxy3D shows high robustness to the streaming case where the inter-frame feature aggregation is undesirable. Even without inter-frame visual relationships, Proxy3D can still establish scene understanding with only instance-based features, which presents a strong contrast to correspondence-based methods that heavily rely on such inter-frame similarities.

\textbf{Dynamic proxy allocation scheme.} In Table~\ref{adaptive_alloc} ablation study we explore the dynamic proxy allocation scheme from Section~\ref{sub:arch} for Scan2Cap benchmark. In this dataset, we initially allocate more proxies to target each object that is being captioned. According to our results, the optimal number of proxies is 5. This is because our adaptive assignment scheme better emphasizes objects of interest in this case, particularly offering more attention to small objects that are usually described with less details. However, as resolution increases (\eg, 5 \vs 10), the proxies for the specific object become less and less informative, causing an imbalance in model understanding across multiple objects.

\section{Conclusion}\label{conc}
In this paper, we have presented the Proxy3D framework with several contributions that further advance spatial intelligence modeling. 
In particular, we have proposed a feature aggregation method that produces compact yet comprehensive proxy representations for 3D scene understanding.  
We have also introduced multi-stage training with iterative development of 3D reasoning skills. 
Finally, our public SpaceSpan dataset with the unified data format has incorporated heterogeneous visual information.
Comprehensive empirical evaluations using various 3D scene understanding tasks have shown competitive or state-of-the-art performance for Proxy3D while using shorter sequences for visual modality.

{
\small
\bibliographystyle{ieeenat_fullname}
\bibliography{main}
}


\end{document}